\documentclass{article}
\usepackage{ijcai21}

\usepackage{times}

\usepackage{soul}
\usepackage{url}
\usepackage[hidelinks]{hyperref}
\usepackage[utf8]{inputenc}
\usepackage[small]{caption}
\usepackage{graphicx}
\usepackage{amsmath}
\usepackage{booktabs}
\usepackage{multirow}
\usepackage{bm}
\usepackage{algorithmic}
\usepackage{subfigure}
\usepackage[ruled, lined, linesnumbered, commentsnumbered, longend]{algorithm2e}
\usepackage{amssymb}

\title{Robust Single-step Adversarial Training with Regularizer}

\author{
	Lehui Xie$^{1,2}$
	\and
	Yaopeng Wang$^{1,2}$\and
	Jia-Li Yin$^{1,2}$\And
	Ximeng Liu$^{1,2}$\\
	\affiliations
	$^1$College of Mathematics and Computer Science, Fuzhou University, Fuzhou 350108, China\\
	$^2$Fujian Provincial Key Laboratory of Information Security of Network Systems, Fuzhou University, Fuzhou 350108, China\\
}

\begin{document}
	
	\maketitle
	
	\begin{abstract}	
		High cost of training time caused by multi-step adversarial example generation is a major challenge in adversarial training. Previous methods try to reduce the computational burden of adversarial training using single-step adversarial example generation schemes, which can effectively improve the efficiency but also introduce the problem of “catastrophic overfitting”, where the robust accuracy against Fast Gradient Sign Method (FGSM) can achieve nearby 100\% whereas the robust accuracy against Projected Gradient Descent (PGD) suddenly drops to 0\% over a single epoch. To address this problem, we propose a novel Fast Gradient Sign Method with PGD Regularization (FGSMPR) to boost the efficiency of adversarial training without catastrophic overfitting. Our core idea is that single-step adversarial training can not learn robust internal representations of FGSM and PGD adversarial examples. Therefore, we design a PGD regularization term to encourage similar embeddings of FGSM and PGD adversarial examples. The experiments demonstrate that our proposed method can train a robust deep network for $L_{\infty}$-perturbations with FGSM adversarial training and reduce the gap to multi-step adversarial training.

	\end{abstract}
	
	\section{Introduction}
	Deep learning has shown outstanding success in near all machine learning fields. However, it has been proved that deep neural networks are vulnerable to adversarial examples, i.e., small disturbances to the input signal, which are usually invisible to the human eyes, is enough to induce large changes in model output \cite{lbfgs}. This phenomenon has aroused people's concerns about the safety of deep learning in the adversarial environment, where malicious attackers may significantly degrade the robustness of deep learning based applications. The vulnerability of deep neural networks has become the focus of extensive attention of researchers.
    
    To mitigate the harm caused by the adversarial attack in deep neural networks, a plethora of defenses have been proposed to train neural networks that are robust to adversarial examples. Several methods \cite{guo2017countering,buckman2018thermometer,metzen2017detecting,33feinman2017detecting,huang2019model,samangouei2018defense} try to detect or pure adversarial examples in the phase of inference. However, \cite{35athalye2018obfuscated} broke a set of purportedly robust defenses, leaving behind adversarial training in which the defender augments each small batch of training data with adversarial examples \cite{31madry2017towards}, one of the few methods that still resists strong attacks.
	
	\begin{figure}[!t]
		\centering
		\includegraphics[scale=0.35]{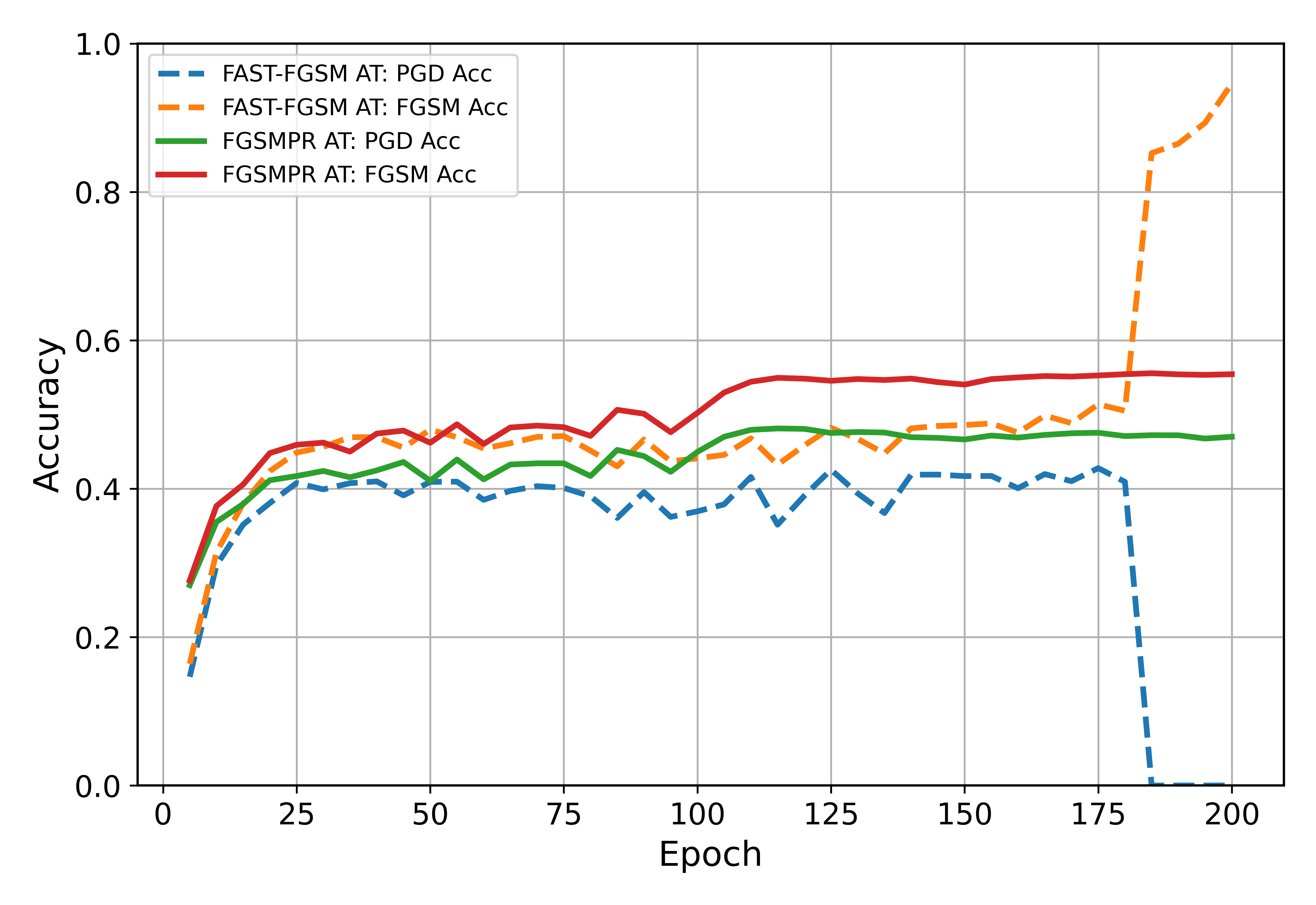}
		\caption{(CIFAR-10) Visualization of the FGSM and PGD robustness of the model trained with FAST-FGSM AT (dashed), FGSMPR AT (solid). All statistics are evaluated against FGSM attacks and 50 steps PGD attacks with 10 random restarts on the test dataset. FAST-FGSM AT occurs catastrophic overfitting at 180 epochs, characterized by a sudden drop of PGD robustness and a rapid increase of FGSM robustness. FGSMPR AT (ours) does not suffer from catastrophic overfitting and maintains stable robustness during the whole training process.}
		\label{merge}
		
	\end{figure}
	
	Adversarial Training (AT) is typically assumed to be more expensive than traditional training due to the necessity of constructing adversarial examples via a first-order method like Projected Gradient Descent (PGD). To combat the increased computational overhead of PGD AT, a recent line of work focused on improving the efficiency of AT. \cite{zhang2019you} proposed to perform multi-step PGD adversarial attacks by chopping off redundant computations during backpropagation when computing adversarial examples to obtain additional speedup. \cite{shafahi2019adversarial} proposed a variant of $K$ steps PGD AT with a single-step Fast Gradient Sign Method (FGSM) AT overhead, called ``FREE AT", which can update model weights as well as input perturbations simultaneously by using a single backpropagation in a way that is less expensive than PGD AT overheads. Inspired by \cite{shafahi2019adversarial}, \cite{wong2020fast} found that previously non-robustness FGSM AT, with a random initialization, could reach similar robustness to PGD AT, called ``FAST-FGSM AT". However, FGSM-based AT suffers from catastrophic overfitting where the robustness against PGD attacks increases in the early stage of training, but suddenly drop to 0 over a single epoch, as shown in Figure \ref{merge}. To address this problem, \cite{andriushchenko2020understanding} proposed Gradient Alignment (GradAlign) to prevent FAST-FGSM AT from suffering catastrophic overfitting. \cite{li2020towards} monitored the FGSM AT process and performed PGD AT with a few batches to help the model recover its robustness when the robustness decreases beyond a threshold. However, these methods are computationally inefficient or fail to overcome catastrophic overfitting.
		
     In this paper, we first analyze the reason why FGSM AT suffers catastrophic overfitting in the training process and show it simply boils down to the fact that FGSM AT is prone to learn spurious functions that excessively fit the FGSM adversarial data distribution but have undefined behavior off the FGSM adversarial data manifold. Then we discuss the difference behind the logits output between the FGSM and PGD adversarial examples in the model trained with FGSM AT and PGD AT, where we show that the logits become significantly different when the FGSM AT trained model suffers from overfitting, while the robust model trained with PGD AT remains stable. We additionally provide for this case an experimental analysis that helps to explain why the FGSM AT trained model generates vastly different logit outputs for single-step and multi-step adversarial examples when catastrophic overfitting occurs. Finally, we propose a novel Fast Gradient Sign Method with PGD regularization (FGSMPR), in which a PGD regularization item is utilized to prompt the model to learn logits that are a function of the truly robust features in the image and ignore the spurious features, thus preventing catastrophic overfitting. 
    
    The contribution of this paper is summarized as follows:
    \begin{itemize}
		\item We analyze the reason why FGSM AT suffers from catastrophic overfitting and demonstrate that the logit distribution of the FGSM AT trained model evaluated against FGSM and PGD adversarial examples becomes significant difference when suffering from catastrophic overfitting.
	
		\item We proposed a Fast Gradient Sign Method with PGD regularization (FGSMPR), which can effectively prevent FGSM AT from catastrophic overfitting by explicitly minimizing the difference in the logit of the model against FGSM and PGD adversarial examples, as shown in Figure \ref{merge}.
		
		\item The extensive experiments show that the FGSMPR can learn a robust model comparable to PGD AT with low computational overhead while does not suffer from catastrophic overfitting. Specially, the FGSMPR takes only 30 minutes to train a CIFAR-10 model with 46\% robustness against 50 steps PGD attacks.
		
	\end{itemize}
	
	\section{Related work and Adversarial Training Overview}
	\subsection{Adversarial Defenses}
	After the concept of the adversarial examples proposed by \cite{lbfgs}. \cite{fgsm} exploited the linearity of deep network models in the higher dimensional space and constructed adversarial examples that use the single sign of gradient as training data to learn robust networks, which is known as FGSM AT. Since the FGSM AT may have a small gradient in the loss function around the original image, the R+FGSM first used random initialization perturbations to step away from the image manifold \cite{59tramer2017ensemble}. The Basic Iterative Method (BIM) \cite{advintherealworld} extended FGSM to iteratively take multiple small steps while adjusting the direction after each step, breaking the FGSM AT. Up to now, the combination of BIM and random initialization \cite{59tramer2017ensemble} has resulted in the well-known PGD AT \cite{31madry2017towards}, which is one of the few methods that can effectively resist to adaptive adversary \cite{35athalye2018obfuscated}. Besides, a large number of defenses include not only AT but also pre-processing or detecting to adversarial examples \cite{guo2017countering,buckman2018thermometer,metzen2017detecting,33feinman2017detecting,huang2019model,samangouei2018defense}. Although the pre-processing techniques or detection algorithms are capable of achieving high defense performance at low computational cost, most of these methods have been broken by \cite{carlini2017adversarial,35athalye2018obfuscated}. Therefore, we focus on AT in this paper.

    \begin{figure*}[!t]
		\centering

		\includegraphics[scale=0.39]{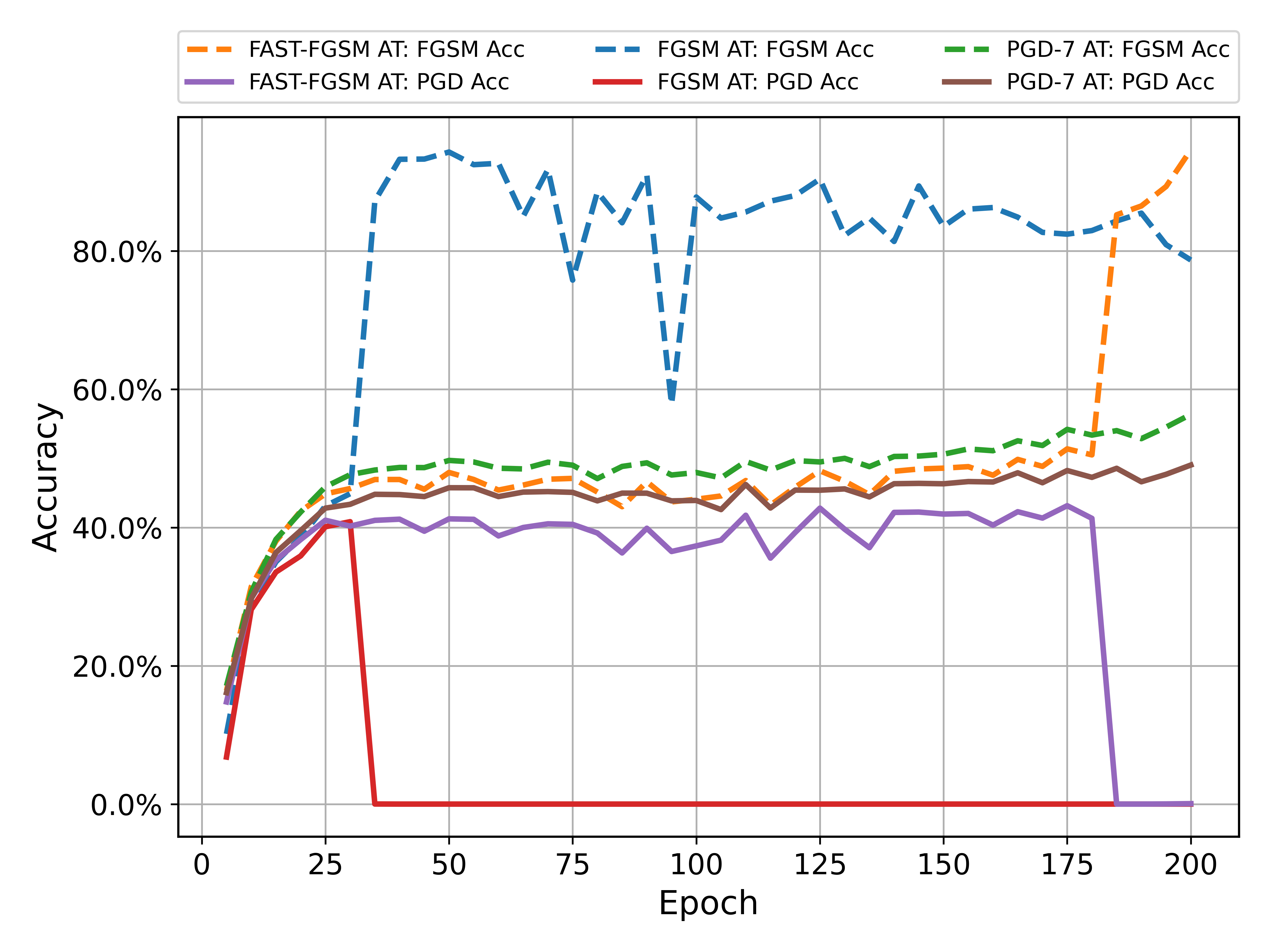}
		\hspace{8mm}
		\includegraphics[scale=0.39]{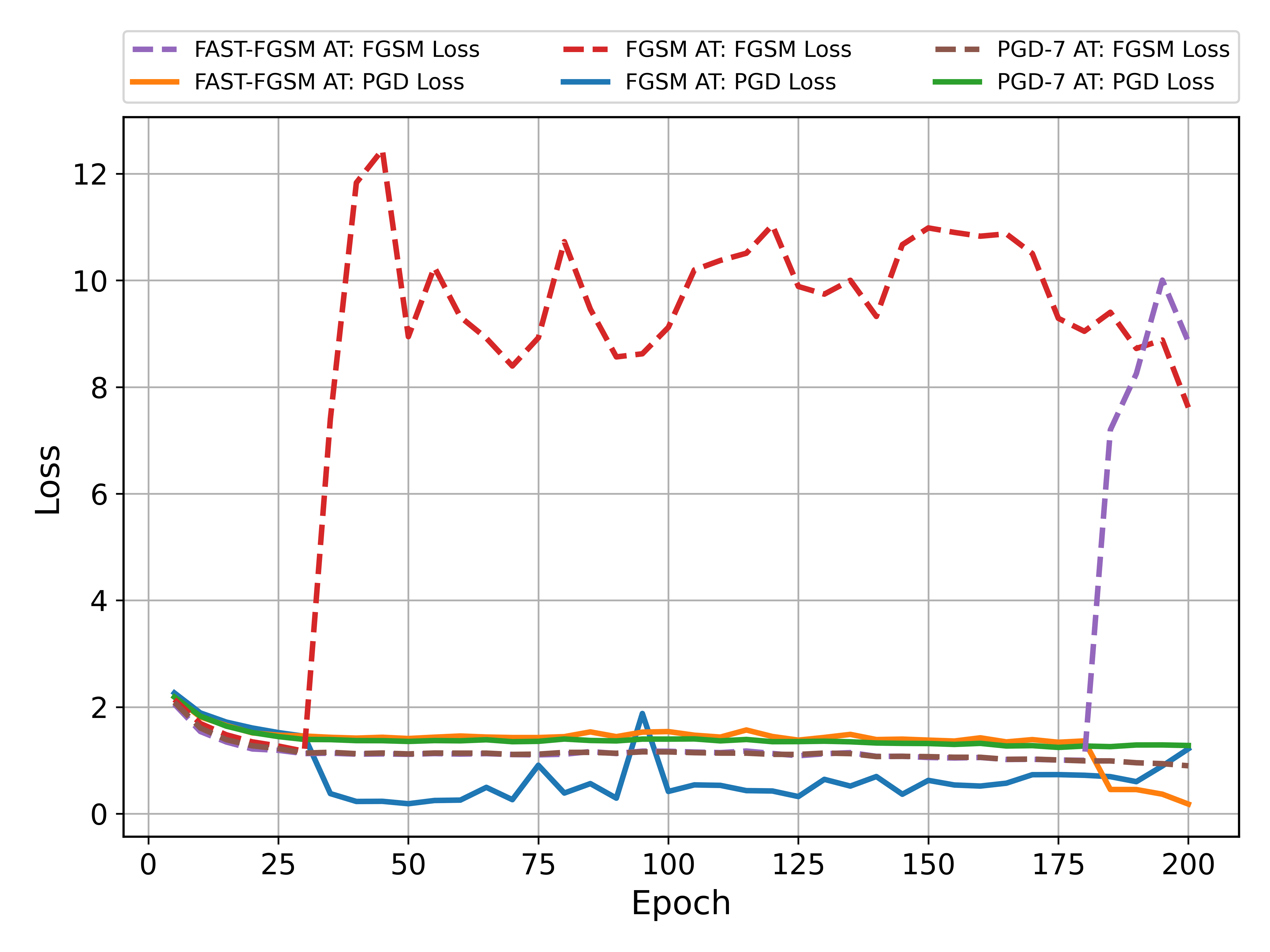}
		
		\caption{(CIFAR-10) Visualization of the FGSM and PGD accuracy/loss of the model trained with FGSM AT, FAST-FGSM AT, PGD-7 AT and tested against FGSM adversarial attacks and 50 steps PGD attack with 10 random restarts during the training process. All results are averaged over three independent runs. FGSM AT and FAST-FGSM AT occurs catastrophic overfitting around 30 and 180 epochs, respectively, characterized by a sudden drop in PGD accuracy and FGSM loss and a rapid increase in PGD loss and FGSM accuracy.}
		\label{diff_epoch_pgd_loss_acc}
	\end{figure*}

	\subsection{Adversarial Training}
	Previous work \cite{31madry2017towards} formalized the training of adversarial robust model into the following non-convex non-concave min-max robust optimization problem:
	\begin{align}\label{eq1}
		\min _{\theta} \mathbb{E}_{(x, y) \sim \mathcal{D}}[\max _{\delta \in \mathcal{S}} \mathcal{L}(\theta, x+\delta, y)].
	\end{align}

	The parameter $\theta$ of the network is learned by Equation \ref{eq1} on the example $(x, y) \sim \mathcal{D}$, where $\mathcal{D}$ is the data generating distribution. $\mathcal{S}$ denotes the region within the $\epsilon$ perturbation range under the $\ell_{\infty}$ threat model for each example, i.e., $\mathcal{S}=\{\delta:\|\delta\|_{\infty} \leq \epsilon\}$, which is usually chosen so that it contains only visually imperceptible perturbations. The procedure for AT is to use adversarial attacks to approximate the internal maximization over $\mathcal{S}$. 
	
	One of the earliest versions of AT used the FGSM attack to find adversarial examples $x'$ to approximate the internal maximization,  formalized as follows \cite{fgsm}:
	\begin{align}
		x' = x + \epsilon \cdot \operatorname{sign}(\nabla_{x} \mathcal{L}(\theta, x, y)).
	\end{align}
	FGSM AT is cheap since it only relies on computing the gradient once. However, the FGSM AT is easily defeated by multi-step adversarial attacks.
	
	PGD attacks \cite{31madry2017towards} used multi-step gradient projection descent to approximate the inner maximization, which is more accurate than FGSM but computationally expensive, formalized as follows:
	\begin{align}
		x^{t+1} & = \Pi_{x+\mathcal{S}}\left(x^{t}+\alpha \operatorname{sign}\left(\nabla_{x} \mathcal{L}(\theta, x, y)\right)\right),
	\end{align}
	where $x^{0}$ initialized as the clean input $x$, $\Pi$ refers to the projection operator, which ensures projecting the adversarial examples back to the ball within the radius $\epsilon$ of the clean data point. The number of iterations $K$ in the PGD attacks (PGD-$K$) determines the strength of the attack and the computational cost. Further, $N$ random restarts are usually employed to verify robustness under strong attacks (PGD-$K$-$N$).
	
	\subsection{Single-step Adversarial Attack for Adversarial Training}
	FREE AT \cite{shafahi2019adversarial}, a single-step training method that generates adversarial examples while updating network weights, is quite similar to FGSM AT. However, FREE AT is robustness against PGD attack that can break the FGSM AT. By deeply analyzing the differences between FREE AT and FGSM AT, \cite{wong2020fast} found that an important property of FREE AT is that the perturbation of the previous sign of gradient is used as the initial perturbation of the next iteration. Based on this observation, \cite{wong2020fast} proposed a FAST-FGSM AT with almost the same robustness as the PGD AT model, but the spent time close to the normal training by adding non-zero initialization perturbations to FGSM AT and further combining some standard techniques \cite{smith2017cyclical,micikevicius2017mixed} to accelerate the model training. Although FAST-FGSM AT largely improves the training speed without sacrificing the robustness of the model, the robustness for PGD adversarial examples suddenly drop to 0\% over a single epoch, which is called catastrophic overfitting. To solve the catastrophic overfitting, \cite{wong2020fast} used the early stopping method to stop training the model when the model robustness decreases beyond a threshold. However, it is obvious that the early stopping cannot determine the moment to stopped so that difficult to balance the performance between robustness and overfitting. 
	
	Recently a series of methods have been proposed to address the catastrophic overfitting problem in single-step AT. \cite{vivek2020single} introduced dropout layers after each non-linear layer of the model and further decay its dropout probability as the training progresses. In addition, \cite{li2020towards} monitored the FGSM AT process and performed PGD AT with a few batches to help the FGSM model recover its robustness when the robustness decreases beyond a threshold. \cite{andriushchenko2020understanding} proposed the Gradient Alignment (GradAlign) regularization item that maximizes the gradient alignment based on the connection between FAST-FGSM AT overfitting and local linearization of the model as a way to prevent the occurrence of catastrophic overfitting. Although these methods provide a better understanding of catastrophic overfitting prevention, but still cannot essentially explain the problem of catastrophic overfitting. Moreover, these methods can improve the robustness of single-step AT models to some extent, but sacrifice a large amount of computational overhead and lose the efficient advantage of single-step AT, even up to the training time of multi-step AT.

	\section{PROPOSED APPROACH}
	\subsection{Observation}\label{ob}
	
	To investigate catastrophic overfitting, we begin by recording the robust accuracy of FGSM AT on CIFAR-10 \cite{CIFAR10}. We evaluate the robust accuracy of the model against 50 steps PGD attacks with 10 random restarts (PGD-50-10) for step size $\alpha= 2/255$ and maximum perturbation $\epsilon=8/255$. Figure \ref{diff_epoch_pgd_loss_acc} visualizes the accuracy and loss of the FGSM AT trained, FAST-FGSM AT trained, and PGD-7 AT trained model and evaluated against FGSM and PGD-50-10 attack during the training phase. As we can see, when FGSM AT and FAST-FGSM AT occur catastrophic overfitting around 30 and 180 epochs respectively, the robustness against PGD-50-10 attack of the model trained with FGSM AT and FAST-FGSM AT begin to drop suddenly, whereas the accuracy against FGSM increases rapidly. However, for the robust PGD-7 AT, the accuracy and loss of the model tend to stabilize after a certain number of epochs.
	
	We maintain that the reason the models trained using FGSM AT suffer from catastrophic overfitting is that it is prone to learn spurious functions that fit the FGSM data distribution but have undefined behavior off the FGSM data manifold. Therefore, the FGSM AT is highly susceptible to overfitting due to a single-step adversarial perturbation, resulting in a sudden drop in the PGD robustness of the model, while the FGSM accuracy increases instantaneously. To study the differences in the performance of the models trained with FGSM AT and PGD-7 AT for evaluating at the FGSM and PGD adversarial examples, we utilize a distance function $\mathcal{L}$ to measure the difference between the output of the model evaluated at single-step and multi-step adversarial attacks. For a model that take inputs $x$ and output logits $f(x)$, we have:
	\begin{align}
		\mathcal{L} (f({x^{fgsm}}), f({x^{pgd}})),
		\label{l2loss}
	\end{align}
    where $x^{fgsm}$ and $x^{pgd}$ are adversarial examples crafted by FGSM and PGD-7, respectively. Here, we choose $L_{2}$ for $\mathcal{L}$. For a well-generalized and robustness model, we assume that the logit $f(x^{fgsm})$ and $f(x^{pgd})$ of the model evaluated at FGSM and PGD adversarial examples should be as similar as possible, i.e., $||f(x^{fgsm})-f(x^{pgd})||_{2}$ should be very small.
	
\begin{figure}[!t]
		\centering
		\subfigure[MNIST]{
			\includegraphics[scale=0.35]{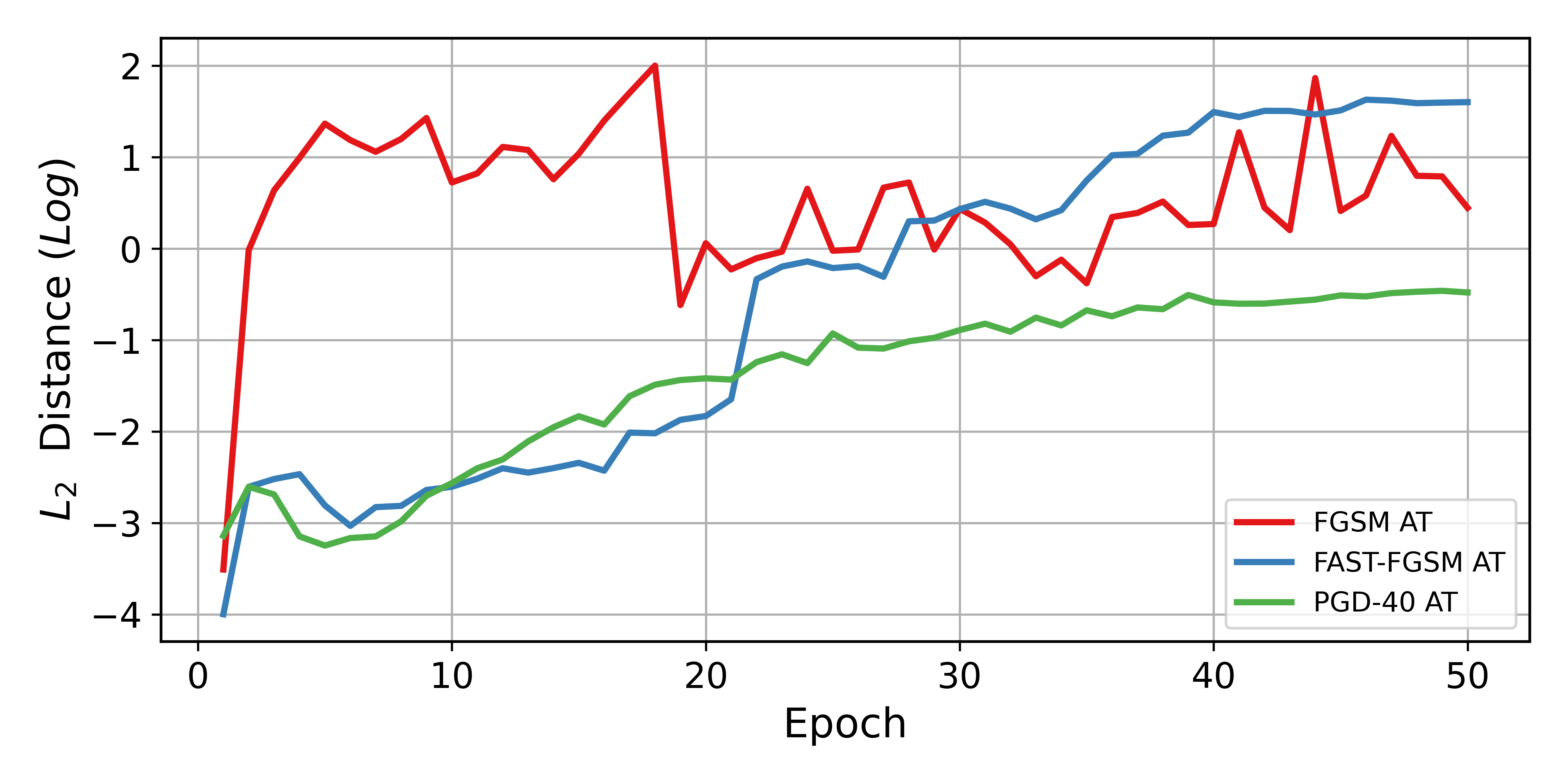}
		}
		\subfigure[CIFAR-10]{
			\includegraphics[scale=0.35]{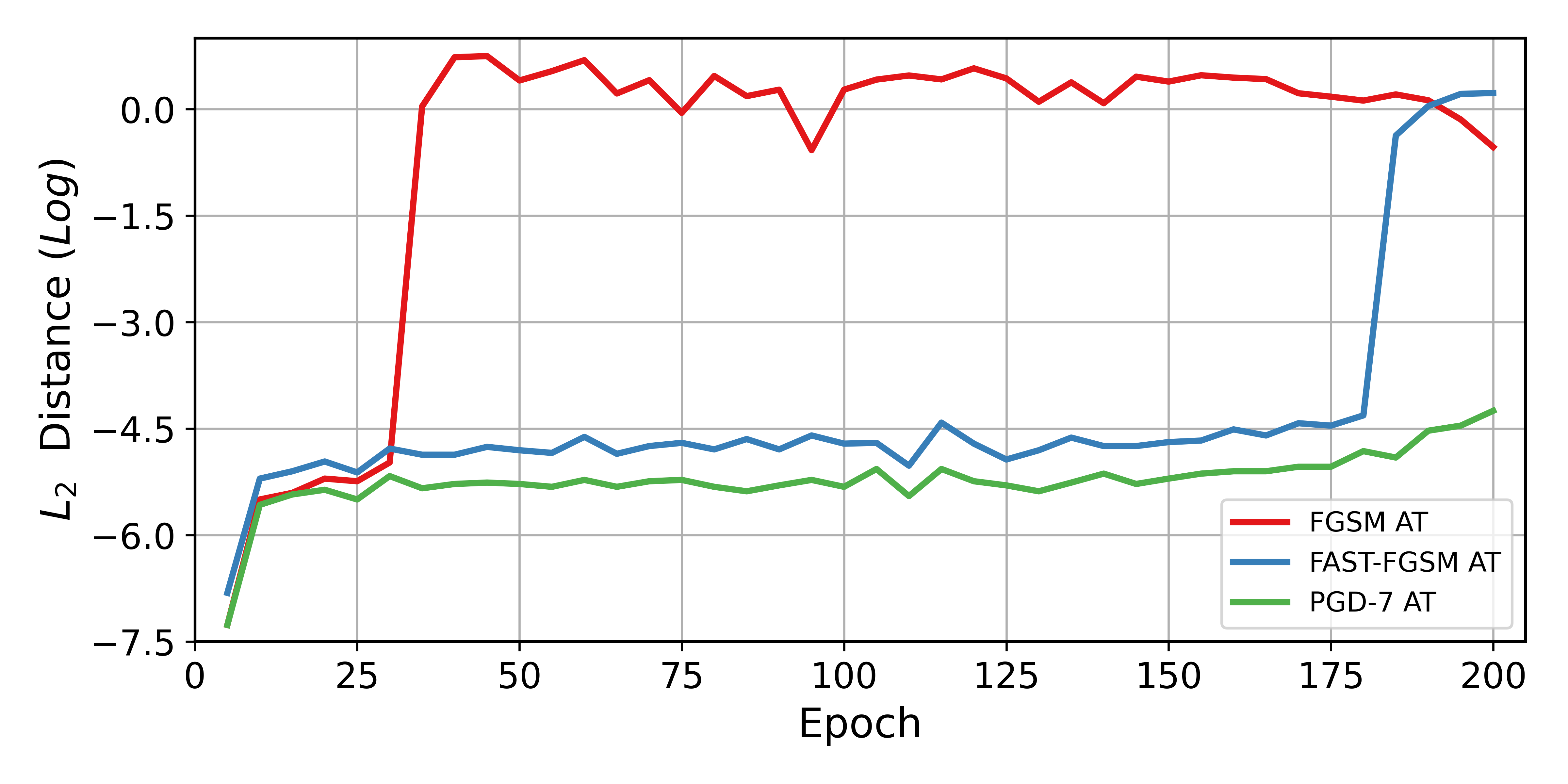}
		}
		\caption{Visualization of the $\mathcal{L}_{2}$ distance of logit of the FGSM AT trained, FAST-FGSM AT trained, PGD-7 AT trained model and evaluated against FGSM and PGD adversarial attack. (a) When the model is not robust, the difference in $\mathcal{L}_{2}$ distance starts to fluctuate, while PGD AT is relatively smooth. (b) FGSM AT and FAST-FGSM AT occurs catastrophic overfitting around 30 and 180 epochs, respectively, and is characterized by a rapid increase of $L_2$ distance.}
		\label{train_on_fgsm_pgd}
	\end{figure}

	To demonstrate our intuition, we firstly train several CIFAR-10 models using FGSM AT and PGD-7 for 200 epochs. For each model, we compute the difference between the output of the model evaluated at FGSM and PGD-7 adversarial examples by using equation \ref{l2loss}, and performed data processing using a logarithmic function to visualize the differences more clearly, as shown in Figure \ref{train_on_fgsm_pgd}. In plot (b), it can be observed that there is no significant difference in the logits from FGSM and PGD adversarial examples during the early phase of training, which matches our intuition. Once catastrophic overfitting occurs, the gap between the logit of the model evaluated at single-step and multi-step adversarial attacks are increasing rapidly around 30 and 180 epochs respectively, which is consistent with PGD loss. In contrast, the PGD-7 AT does not suffer catastrophic overfitting and the difference of the logit of the model is keeping stable.	This phenomenon will also appear on the simple MNIST dataset \cite{MNIST}, but it is not as clear as CIFAR-10, as shown in plot (a).

	\subsection{PGD Regularization}

	Based on the analysis in Section \ref{ob}, the only FGSM adversarial loss is not enough for the model to learn the robust features of both single-step and multi-step adversarial examples. To solve this problem, inspired by \cite{58kannan2018adversarial}, we use the logit pairing to encourage the model to learn robust internal representation of FGSM and PGD adversarial examples so that the logit outputs $f(x^{fgsm})$ and $f(x^{pgd})$ of the model for FGSM and PGD adversarial examples to be as similar as possible:
	\begin{align}{\label{reg_loss}}
		\lambda \frac{1}{m} \sum_{i = 1}^{m} \mathcal{L}(f(x_{i}^{fgsm} ; \theta), f({x}_{i}^{pgd};\theta)),
	\end{align}
	where $\mathcal{L}$ is $L_{2}$ norm; $x_{i}^{fgsm}$ and ${x}_{i}^{pgd}$ are adversarial examples crafted by FGSM and PGD attacks, respectively; $\lambda$ is a hyparameter to balance FGSM loss and PGD regularization item. Combining with the proposed regularization, the FGSM AT can learn a robustness model comparable with PGD-7 AT, as validated in Section \ref{experiment}. 
	
	We hold that PGD regularization works well because it provides an additional prior that regularizes the model toward a more accurate understanding of adversarial examples. If we train the model with only the single-step FGSM adversarial loss, it is prone to learn spurious functions that excessively fit the FGSM adversarial data distribution but have undefined behavior off the FGSM data manifold (e.g., multi-step adversarial examples). PGD regularization forces the explanations of the FGSM adversarial example and multi-step adversarial example to be similar. This is essentially a prior encouraging the model to learn logits that are a function of the truly significant features in the image and ignore the spurious features.
	
	\subsection{Training Route}
	The overall training procedure of the FGSMPR AT is summarized in Algorithm \ref{trianing}. We first perform FGSM adversarial attack to generate FGSM adversarial examples $x^{fgsm}_{i}$ and compute FGSM AT loss $fgsm\_loss$ using cross-entropy. Then, we perform PGD adversarial attack for $m$ examples, from a batch of natural examples, to generate $m$ PGD adversarial examples. After generating FGSM and PGD adversarial examples, the regularization loss $reg\_loss$ of $m$ FGSM and PGD adversarial examples are calculated using Equation \ref{reg_loss} and used as part of the total loss $total\_loss$. Finally, the parameter $\theta$ of the model is updated using a proper optimizer (e.g., stochastic gradient descent). The hyperparameter $\lambda$ shall be properly chosen to balance FGSM loss $fgsm\_loss$ and PGD regularization $reg\_loss$ item. In practice, we take $\alpha=\epsilon/K$, $K = 3$ and $m = 1$. In other words, we only pick a single example from a batch for generating a PGD-3 adversarial example, which is then used for regularization to encourage the model to learn similar logit output. The experiments show that a single PGD adversarial example for regularization is enough to learn a robustness model.
	
	\begin{algorithm}[!t]
		\SetKwFunction{isOddNumber}{isOddNumber}
		\SetKwInOut{KwIn}{Input}
		\SetKwInOut{KwOut}{Output}
		\KwIn{
			Training data $(X, Y)$, perturbation bound $\epsilon$, learning rate $\gamma$, hyparameter $\alpha, \lambda$. 
		}
		\KwOut{Trained model $f(\cdot)$ with parameter $\theta$}
		\For{epoch = 1 ... $N_{epoch}$ }{
			\For{i = 1 ... $B$ }{
			    // Perform FGSM adversarial attack \\
				$x_{i}^{fgsm} = x_{i} + \alpha \cdot \operatorname{sign}(\nabla_{\delta} \ell(f_{\theta}(x_{i}), y_{i}))$\\
				$fgsm\_loss = J({x_{i}^{fgsm}},\theta)$	\\
				
				// Perform PGD adversarial attack \\
				\For{k = 1 ... $K$ }{
					$\delta = \delta + \alpha \cdot \operatorname{sign}(\nabla_{\delta} \ell(f_{\theta}(x_{i}+\delta), y_{i}))$ \\
					$\delta = \max(\min(\delta, \epsilon), -\epsilon))$
				}
				
				$x_{i}^{pgd} = x_{i} + \delta$ \\
				$reg\_loss = \lambda \frac{1}{m} \sum_{j = 1}^{m} \mathcal{L}(f(x_{i,j}^{fgsm} ; \theta), f({x}_{i,j}^{pgd};\theta))$	\\
				$total\_loss = fgsm\_loss + reg\_loss$ \\
				Update model parameter $\theta$ based on $total\_loss$}
		}
		\KwRet{$f(\cdot)$.}
		\caption{FGSMPR AT}
		\label{trianing}
	
	\end{algorithm}

	\section{Experiments}\label{experiment}
	In this section, we demonstrate that the proposed FGSMPR is robust against strong PGD attacks. All experiments are run on a single RTX 2070, in which we use half-precision computation recommended in \cite{wong2020fast} to speed up the training of CIFAR-10 model, which was incorporated with the Apes amp package at the O1 optimization level for all CIFAR-10 experiments.
	\newline
	\textbf{Attacks:} We attack all models using PGD attacks with $K$ iterations and 10 random restarts on both cross-entropy loss (PGD-K-10) and the Carlini-Wagner loss (CW-K-10) \cite{cwattack}. All PGD attacks used at evaluation for MNIST \cite{MNIST} are run with 10 random restarts for 20/40 iterations. All PGD attacks used at evaluation for CIFAR-10 \cite{CIFAR10} are run with 10 random restarts for 20/50 steps.
	\newline
	\textbf{Perturbation:} For MNIST, we set the maximum perturbation $\epsilon$ to 0.3 and the PGD step size $\alpha$ to 0.1. For CIFAR-10, we set the maximum perturbation $\epsilon$ to $8/255$ and the PGD step size $\alpha$ to $2/255$.
	\newline
	\textbf{Comparisions:} 	We compare the performance of our proposed method (FGSMPR) with FGSM: standard FGSM AT \cite{fgsm}; FAST-FGSM AT: FGSM AT with a random initialization \cite{wong2020fast}; FREE AT: recently proposed single-step AT method \cite{shafahi2019adversarial}; GradAlign AT: recently proposed method solving catastrophic overfitting \cite{andriushchenko2020understanding}; PGD-$K$ AT: AT with a $K$ iterations PGD attack \cite{31madry2017towards}. 
	\newline
	\textbf{Evaluation:} We demonstrate that the performance of models against PGD-$K$-10/CW-$K$-10 adversarial attacks under white-box settings. For all experiments, the mean and standard deviation over three independent runs are reported.

	\subsection{Results on MNIST}
	First, we conduct a study to demonstrate that our proposed approach is highly working in MNIST benchmark dataset \cite{MNIST}. We train models for MNIST dataset with the same architecture used by \cite{wong2020fast}, using FGSM AT, FAST-FGSM AT, FREE AT, PGD-40 AT, FGSMPR AT. Except that the AT free replays each batch of $m=8$ for a total of 7 epochs, all other models are trained for 50 epochs. For the proposed method, we set the hyparameter $\lambda$, $K$ and $m$ to $(0.1, 3, 1)$. The experimental results are provided in Table \ref{mnist}. It can be observed that our proposed FGSMPR AT is more robust against both PGD and CW attacks on the MNIST dataset than the GradAlign AT and FREE AT, and is second only to the PGD AT model with a small difference. In the course of testing the robustness of FAST-FGSM AT on the MNIST dataset, we found an interesting problem where increasing the number of MNIST training epochs to 50 also resulted in catastrophic overfitting, although this phenomenon was previously found only in CIFAR-10. Besides, the GradAlign AT \cite{andriushchenko2020understanding} can keep the model from suffering catastrophic overfitting to some extent, but it is far inferior to other comparison methods in defending against the higher iteration adversarial attacks.

    \begin{table*}[!t]
        \centering
        \caption{Validation accuracy (\%) and robustness of MNIST models trained with FGSM AT, FAST-FGSM AT, GradAlign AT, FREE AT, PGD-40 AT, FGSMPR AT without early stopping and the corresponding training time. All statistics are evaluated against PGD/CW attacks with 20/40 iterations and 10 random restarts for $\alpha=0.1$, $\epsilon=0.3$ over three independent runs.}
        \label{mnist}
        \resizebox{0.98\textwidth}{!}{%
        \begin{tabular}{lrrrrrr}
        
        \textbf{Method} & \textbf{Standard Accuracy} & \textbf{PGD-20-10} & \textbf{PGD-40-10} & \textbf{CW-20-10} & \textbf{CW-40-10} & \textbf{Training Time (s)} \\ \hline
        FGSM AT          & 97.53$\pm$0.39 & 39.31$\pm$20.68 & 12.45$\pm$12.15 & 40.14$\pm$20.64 & 13.46$\pm$12.89 & 481.03$\pm$0.81  \\
        FAST-FGSM AT     & 98.52$\pm$0.34 & 42.60$\pm$12.47 & 11.19$\pm$6.52  & 43.41$\pm$12.55 & 11.85$\pm$7.26  & 491.24$\pm$1.71  \\
        GradAlign AT     & 99.05$\pm$0.03 & 91.42$\pm$0.57  & 75.94$\pm$3.23  & 91.23$\pm$0.51  & 75.86$\pm$3.16  & 633.96$\pm$3.55  \\
        FREE AT          & 98.49$\pm$0.05 & 92.90$\pm$0.20  & 90.06$\pm$0.36  & 92.70$\pm$0.15  & 89.85$\pm$0.32  & 175.45$\pm$1.99  \\
        PGD-40 AT        & 99.16$\pm$0.03 & 94.72$\pm$0.08  & 92.52$\pm$0.14  & 94.75$\pm$0.03  & 92.65$\pm$0.10  & 3652.39$\pm$1.00 \\
        FGSMPR AT (ours) & 98.35$\pm$0.09 & 93.77$\pm$0.32  & 90.83$\pm$0.49  & 93.65$\pm$0.26  & 90.56$\pm$0.55  & 626.57$\pm$1.68  \\ \hline
        \end{tabular}%
        }
    \end{table*}

    \begin{table*}[!t]
    \centering
        \caption{Validation accuracy (\%) and robustness of CIFAR-10 models trained with FGSM AT, FAST-FGSM AT, GradAlign AT, FREE AT, PGD-7 AT, FGSMPR AT without early stopping and the corresponding training time. All statistics are evaluated against PGD/CW attacks with 20/50 iterations and 10 random restarts for $\alpha=2/255$, $\epsilon=8/255$ over three independent runs.}
        \label{cifar10}
        \resizebox{0.98\textwidth}{!}{%
        \begin{tabular}{lrrrrrr}
  
        \textbf{Method} & \textbf{Standard Accuracy} & \textbf{PGD-20-10} & \textbf{PGD-50-10} & \textbf{CW-20-10} & \textbf{CW-50-10} & \textbf{Training Time (m)} \\ \hline
        FGSM AT          & 88.51$\pm$1.27 & 0.01$\pm$0.17  & 0.00$\pm$0.00  & 0.01$\pm$0.11  & 0.00$\pm$0.00  & 119.04$\pm$0.41 \\
        FAST-FGSM AT     & 90.33$\pm$0.42 & 0.92$\pm$0.49  & 0.32$\pm$0.25  & 0.52$\pm$0.30  & 00.17$\pm$0.08 & 123.81$\pm$0.18 \\
        GradAlign AT     & 82.82$\pm$0.13 & 32.94$\pm$0.83 & 32.50$\pm$0.80 & 32.94$\pm$0.83 & 32.52$\pm$0.81 & 486.20$\pm$0.67 \\
        FREE AT          & 82.32$\pm$0.12 & 46.97$\pm$0.05 & 46.07$\pm$0.82 & 45.77$\pm$0.23 & 45.64$\pm$0.24 & 61.94$\pm$0.15  \\
        PGD-7 AT         & 84.75$\pm$0.87 & 48.33$\pm$0.62 & 47.99$\pm$0.66 & 47.80$\pm$0.31 & 47.59$\pm$0.38 & 493.41$\pm$0.04 \\ 
        FGSMPR AT (ours) & 83.31$\pm$0.40 & 47.59$\pm$0.51 & 47.19$\pm$0.42 & 46.98$\pm$0.18 & 46.79$\pm$0.20 & 211.65$\pm$0.61 \\ \hline
        \end{tabular}%
        }
    \end{table*}

	\subsection{Results on CIFAR-10}
\begin{figure}[!t]
		\centering
		\includegraphics[scale=0.4]{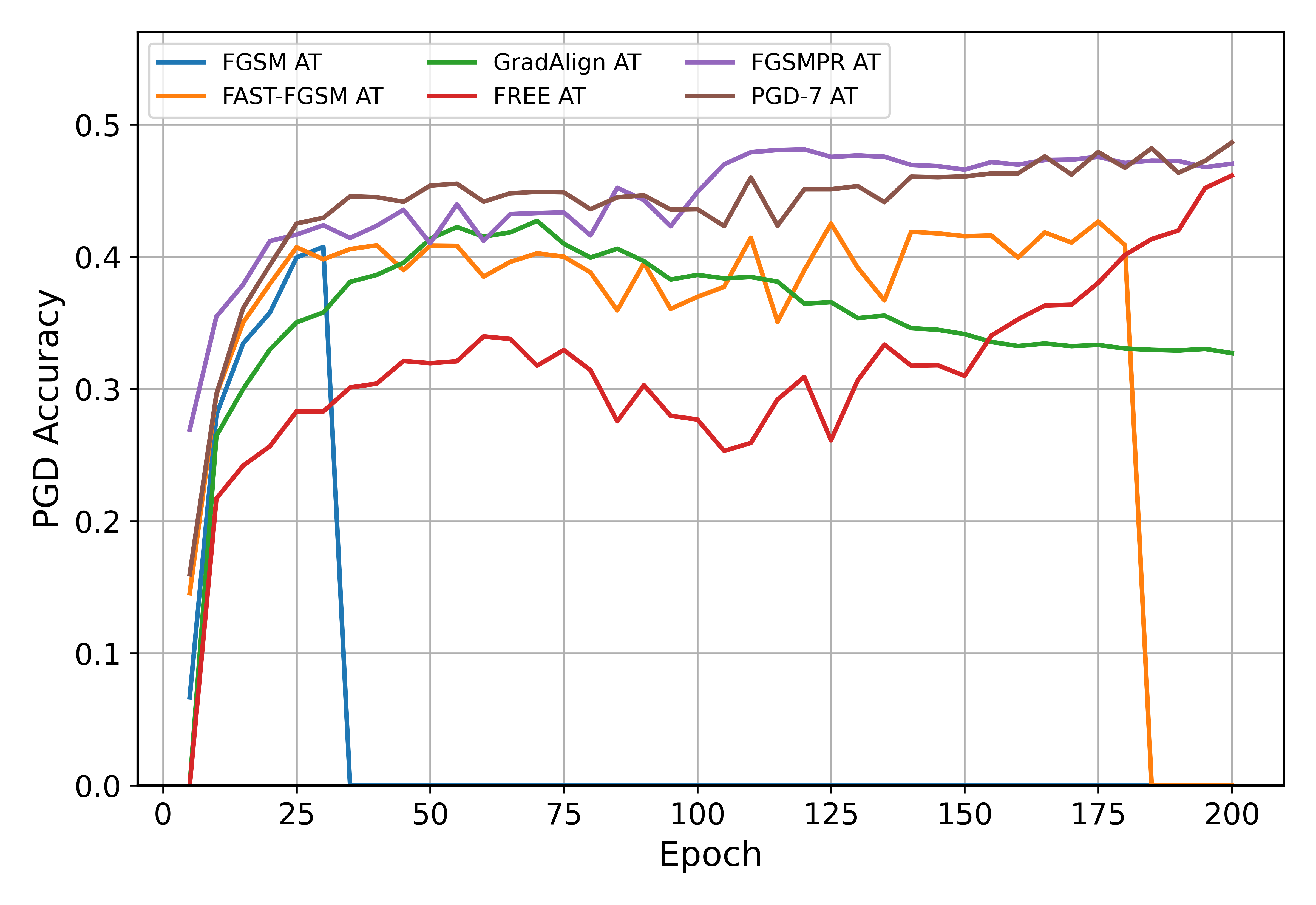}
		\caption{Visualization of the accuracy of the CIFAR-10 model trained for FGSM AT, FAST-FGSM AT, GradAlign AT, FREE AT, PGD-7 AT, and FGSMPR AT. All the statistics are tested against 50 steps PGD attacks with 10 random restarts for $\alpha=2/255$, $\epsilon=8/255$. Catastrophic overfitting for the FGSM and FAST-FGSM AT occur around 30 and 180 epochs, respectively, and is characterized by a sudden drop in the PGD accuracy.}
		\label{diff_epoch_pgd_acc}
	\end{figure}

\begin{figure}[!t]
		\centering
		\includegraphics[scale=0.4]{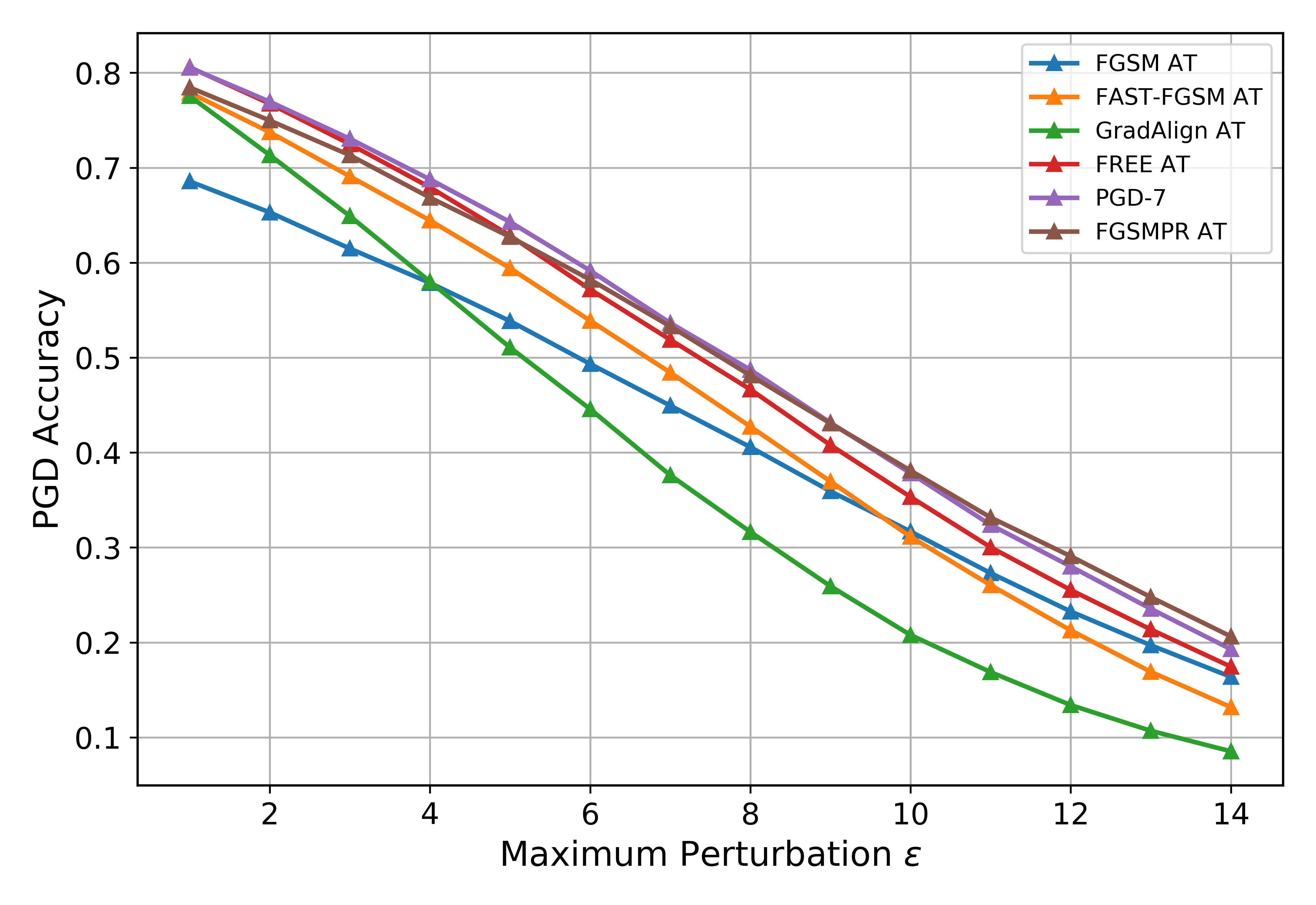}
		\caption{Accuracy of the model trained for FGSM AT, FAST-FGSM AT, GradAlign AT, FREE AT, PGD-7 AT and FGSMPR AT with early stopping. All the statistics are evaluated against 50 steps PGD attacks with 10 random restarts for $l_{\infty}$-perturbation $\epsilon$.}
		\label{diff_eps_cmp}
	\end{figure}

	To verify whether AT scheme suffers from catastrophic overfitting, we train 200 epoch for all CIFAR-10 models using the Preact ResNet-18 \cite{he2016identity} architecture without early stopping, especially the FREE AT replays each batch $m=8$ times for a total of 25 epochs as recommend in \cite{shafahi2019adversarial}. For the FGSMPR, we set the hyparameter $\lambda$, $K$ and $m$ to $(0.5, 3, 1)$. The experimental results are provided in Table \ref{cifar10}. It can be observed that FGSMPR AT is quite similar to PGD-7 AT while our training time is half of PGD-7 AT. To demonstrate that the proposed FGSMPR does not suffer from catastrophic overfitting, we takes 211 minutes to train a CIFAR-10 model for 200 epochs, which is longer than time for FREE AT. However, our method was able to achieve 46\% robustness by training 30 epochs in only 30 minutes, which is half less than FREE AT. Further, we visualize the robustness of the training process of different AT method and tested against a 10 random restart PGD-50 attack, as shown in Figure \ref{diff_epoch_pgd_acc}. It can be observed that the robustness of the FGSMPR against PGD has steadily increased, which is only 0.8\% behind PGD-7 AT and does not suffer from catastrophic overfitting even when trained to 200 epochs. Instead, FAST-FGSM AT started to have a trend similar to PGD AT, but there is a sharp drop in robustness around 180 epochs when occuring catastrophic overfitting. The GradAlign AT was proposed to prevent the FGSM AT from catastrophic overfitting, but the accuracy still dropped by more than 10\% and took more than two times longer compared to our FGSMPR AT. Besides, we also evaluate robustness of the model under different $l_{\infty}$ perturbation where all models are trained with early stopping. In the case of larger $l_{\infty}$ perturbations, FGSMPR AT is essentially indistinguishable from PGD-7 AT, and even slightly better than PGD-7 AT, as shown in Figure \ref{diff_eps_cmp}.
	
	\section{Conclusion}
	In this paper, we analyze the reason for FGSM AT suffers catastrophic overfitting and show that FGSM AT is prone to learn spurious functions that fit the FGSM adversarial data distribution but have undefined behavior off the FGSM data manifold. Further, we discuss the difference behind the logits between the FGSM and PGD adversarial examples in the model trained with FGSM AT and PGD AT, where the logit becomes significantly different when FGSM AT suffers from overfitting, while PGD AT remains stable. Therefore, we propose a novel FGSMPR AT, where a PGD regularization term is used to encourage the model to learn similar embeddings of FGSM and PGD adversarial examples. The extensive experiments show that the FGSMPR can effectively keep FGSM AT from catastrophic overfitting with a low computational cost.
	
    \clearpage
	\bibliographystyle{named}

\begin{thebibliography}{}
		
		\bibitem[\protect\citeauthoryear{Andriushchenko and
			Flammarion}{2020}]{andriushchenko2020understanding}
		Maksym Andriushchenko and Nicolas Flammarion.
		\newblock Understanding and improving fast adversarial training.
		\newblock {\em arXiv preprint arXiv:2007.02617}, 2020.
		
		\bibitem[\protect\citeauthoryear{Athalye \bgroup \em et al.\egroup
		}{2018}]{35athalye2018obfuscated}
		Anish Athalye, Nicholas Carlini, and David Wagner.
		\newblock Obfuscated gradients give a false sense of security: Circumventing
		defenses to adversarial examples, 2018.
		
		\bibitem[\protect\citeauthoryear{Buckman \bgroup \em et al.\egroup
		}{2018}]{buckman2018thermometer}
		Jacob Buckman, Aurko Roy, Colin Raffel, and Ian Goodfellow.
		\newblock Thermometer encoding: One hot way to resist adversarial examples.
		\newblock In {\em International Conference on Learning Representations}, 2018.
		
		\bibitem[\protect\citeauthoryear{Carlini and
			Wagner}{2017a}]{carlini2017adversarial}
		Nicholas Carlini and David Wagner.
		\newblock Adversarial examples are not easily detected: Bypassing ten detection
		methods.
		\newblock In {\em Proceedings of the 10th ACM Workshop on Artificial
			Intelligence and Security}, pages 3--14, 2017.
		
		\bibitem[\protect\citeauthoryear{Carlini and Wagner}{2017b}]{cwattack}
		Nicholas Carlini and David Wagner.
		\newblock Towards evaluating the robustness of neural networks.
		\newblock In {\em 2017 ieee symposium on security and privacy (sp)}, pages
		39--57. IEEE, 2017.
		
		\bibitem[\protect\citeauthoryear{Feinman \bgroup \em et al.\egroup
		}{2017}]{33feinman2017detecting}
		Reuben Feinman, Ryan~R Curtin, Saurabh Shintre, and Andrew~B Gardner.
		\newblock Detecting adversarial samples from artifacts.
		\newblock {\em arXiv preprint arXiv:1703.00410}, 2017.
		
		\bibitem[\protect\citeauthoryear{Goodfellow \bgroup \em et al.\egroup
		}{2014}]{fgsm}
		Ian~J. Goodfellow, Jonathon Shlens, and Christian Szegedy.
		\newblock Explaining and harnessing adversarial examples, 2014.
		
		\bibitem[\protect\citeauthoryear{Guo \bgroup \em et al.\egroup
		}{2017}]{guo2017countering}
		Chuan Guo, Mayank Rana, Moustapha Cisse, and Laurens Van Der~Maaten.
		\newblock Countering adversarial images using input transformations.
		\newblock {\em arXiv preprint arXiv:1711.00117}, 2017.
		
		\bibitem[\protect\citeauthoryear{He \bgroup \em et al.\egroup
		}{2016}]{he2016identity}
		Kaiming He, Xiangyu Zhang, Shaoqing Ren, and Jian Sun.
		\newblock Identity mappings in deep residual networks.
		\newblock In {\em European conference on computer vision}, pages 630--645.
		Springer, 2016.
		
		\bibitem[\protect\citeauthoryear{Huang \bgroup \em et al.\egroup
		}{2019}]{huang2019model}
		Bo~Huang, Yi~Wang, and Wei Wang.
		\newblock Model-agnostic adversarial detection by random perturbations.
		\newblock In {\em IJCAI}, pages 4689--4696, 2019.
		
		\bibitem[\protect\citeauthoryear{Kannan \bgroup \em et al.\egroup
		}{2018}]{58kannan2018adversarial}
		Harini Kannan, Alexey Kurakin, and Ian Goodfellow.
		\newblock Adversarial logit pairing, 2018.
		
		\bibitem[\protect\citeauthoryear{Krizhevsky \bgroup \em et al.\egroup
		}{2009}]{CIFAR10}
		Alex Krizhevsky, Geoffrey Hinton, et~al.
		\newblock Learning multiple layers of features from tiny images.
		\newblock https://www.cs.toronto.edu/~kriz/learning-features-2009-TR.pdf, 2009.
		
		\bibitem[\protect\citeauthoryear{Kurakin \bgroup \em et al.\egroup
		}{2016}]{advintherealworld}
		Alexey Kurakin, Ian Goodfellow, and Samy Bengio.
		\newblock Adversarial examples in the physical world.
		\newblock {\em arXiv preprint arXiv:1607.02533}, 2016.
		
		\bibitem[\protect\citeauthoryear{LeCun \bgroup \em et al.\egroup
		}{1998}]{MNIST}
		Yann LeCun, L{\'e}on Bottou, Yoshua Bengio, and Patrick Haffner.
		\newblock Gradient-based learning applied to document recognition.
		\newblock {\em Proceedings of the IEEE}, 86(11):2278--2324, 1998.
		
		\bibitem[\protect\citeauthoryear{Li \bgroup \em et al.\egroup
		}{2020}]{li2020towards}
		Bai Li, Shiqi Wang, Suman Jana, and Lawrence Carin.
		\newblock Towards understanding fast adversarial training.
		\newblock {\em arXiv preprint arXiv:2006.03089}, 2020.
		
		\bibitem[\protect\citeauthoryear{Madry \bgroup \em et al.\egroup
		}{2017}]{31madry2017towards}
		Aleksander Madry, Aleksandar Makelov, Ludwig Schmidt, Dimitris Tsipras, and
		Adrian Vladu.
		\newblock Towards deep learning models resistant to adversarial attacks, 2017.
		
		\bibitem[\protect\citeauthoryear{Metzen \bgroup \em et al.\egroup
		}{2017}]{metzen2017detecting}
		Jan~Hendrik Metzen, Tim Genewein, Volker Fischer, and Bastian Bischoff.
		\newblock On detecting adversarial perturbations.
		\newblock {\em arXiv preprint arXiv:1702.04267}, 2017.
		
		\bibitem[\protect\citeauthoryear{Micikevicius \bgroup \em et al.\egroup
		}{2017}]{micikevicius2017mixed}
		Paulius Micikevicius, Sharan Narang, Jonah Alben, Gregory Diamos, Erich Elsen,
		David Garcia, Boris Ginsburg, Michael Houston, Oleksii Kuchaiev, Ganesh
		Venkatesh, et~al.
		\newblock Mixed precision training.
		\newblock {\em arXiv preprint arXiv:1710.03740}, 2017.
		
		\bibitem[\protect\citeauthoryear{Samangouei \bgroup \em et al.\egroup
		}{2018}]{samangouei2018defense}
		Pouya Samangouei, Maya Kabkab, and Rama Chellappa.
		\newblock Defense-gan: Protecting classifiers against adversarial attacks using
		generative models.
		\newblock {\em arXiv preprint arXiv:1805.06605}, 2018.
		
		\bibitem[\protect\citeauthoryear{Shafahi \bgroup \em et al.\egroup
		}{2019}]{shafahi2019adversarial}
		Ali Shafahi, Mahyar Najibi, Mohammad~Amin Ghiasi, Zheng Xu, John Dickerson,
		Christoph Studer, Larry~S Davis, Gavin Taylor, and Tom Goldstein.
		\newblock Adversarial training for free!
		\newblock In {\em Advances in Neural Information Processing Systems}, pages
		3358--3369, 2019.
		
		\bibitem[\protect\citeauthoryear{Smith}{2017}]{smith2017cyclical}
		Leslie~N Smith.
		\newblock Cyclical learning rates for training neural networks.
		\newblock In {\em 2017 IEEE Winter Conference on Applications of Computer
			Vision (WACV)}, pages 464--472. IEEE, 2017.
		
		\bibitem[\protect\citeauthoryear{Szegedy \bgroup \em et al.\egroup
		}{2013}]{lbfgs}
		Christian Szegedy, Wojciech Zaremba, Ilya Sutskever, Joan Bruna, Dumitru Erhan,
		Ian Goodfellow, and Rob Fergus.
		\newblock Intriguing properties of neural networks, 2013.
		
		\bibitem[\protect\citeauthoryear{Tramèr \bgroup \em et al.\egroup
		}{2017}]{59tramer2017ensemble}
		Florian Tramèr, Alexey Kurakin, Nicolas Papernot, Ian Goodfellow, Dan Boneh,
		and Patrick McDaniel.
		\newblock Ensemble adversarial training: Attacks and defenses, 2017.
		
		\bibitem[\protect\citeauthoryear{Vivek and Babu}{2020}]{vivek2020single}
		BS~Vivek and R~Venkatesh Babu.
		\newblock Single-step adversarial training with dropout scheduling.
		\newblock In {\em 2020 IEEE/CVF Conference on Computer Vision and Pattern
			Recognition (CVPR)}, pages 947--956. IEEE, 2020.
		
		\bibitem[\protect\citeauthoryear{Wong \bgroup \em et al.\egroup
		}{2020}]{wong2020fast}
		Eric Wong, Leslie Rice, and J~Zico Kolter.
		\newblock Fast is better than free: Revisiting adversarial training.
		\newblock {\em arXiv preprint arXiv:2001.03994}, 2020.
		
		\bibitem[\protect\citeauthoryear{Zhang \bgroup \em et al.\egroup
		}{2019}]{zhang2019you}
		Dinghuai Zhang, Tianyuan Zhang, Yiping Lu, Zhanxing Zhu, and Bin Dong.
		\newblock You only propagate once: Accelerating adversarial training via
		maximal principle.
		\newblock In {\em Advances in Neural Information Processing Systems}, pages
		227--238, 2019.
		
	\end{thebibliography}

\end{document}